# Utilizing UNet for the future traffic map prediction task: Traffic4cast challenge 2020'

Sungbin Choi

sungbin.choi.1@gmail.com

**Abstract.** This paper describes our UNet based experiments on the Traffic4cast challenge 2020'. Similar to the Traffic4cast challenge 2019, the task is to predict traffic flow volume, direction and speed on a high resolution map of three large cities worldwide. We mainly experimented with UNet based deep convolutional networks with various compositions of densely connected convolution layers, average pooling layers and max pooling layers. Three base UNet model types are tried and predictions are combined by averaging prediction scores or taking median value. Our method achieved best performance in this year's newly built challenge dataset.

## 1. Introduction

This is our second participation on this Traffic4cast (Traffic map movie forecasting) challenge [1] following Traffic4cast 2019' [2, 3]. Challenge's task is to predict future traffic flow volume, heading and speed on a high resolution map of three large cities worldwide. Contrary to last year's challenge task which was to predict the next fifteen minutes at maximum time distance, this year's task gets more challenging because we need to predict more distant future traffic maps up to one hour later.

In addition to our previous works on UNet based approach on this traffic prediction task, we tried to experiment with a more diverse set of neural net structures and combined them to improve performance beyond any single model can achieve.

## 2. Methods

**2.1 Task**

All traffic images have 495 x 436 image dimension (height and width). Each pixel represents a 100 square meter area. It is captured per 5 minute time interval. Given a 12 timeframe traffic map image which represents one hour long traffic map of cities, we need to predict the future 6 timeframe, which represents the next 5, 10, 15, 30, 45, 60 minute timeframe traffic maps respectively.

**2.2 Input**

Challenge's input data is composed of dynamic input data which varies over time, and static input data which is invariant to the time.

*2.2.1 Dynamic input data*

Dynamic input data is (12, 495, 436, 9) shaped tensor. First axis represents the time bin and the last axis represents the feature channel. 9 feature channel contains traffic volume and speed for four headings (northeast, northwest, southeast and southwest) respectively, plus computed incident level. Each channel is given normalized and discretized to range from minimum 0 to maximum 255.

In this study, we combined the time bin axis with the feature channel, so input data is converted to (495, 436, 108) shaped tensor.

*2.2.2 Static input data*

Static input data is (495, 436, 7) shaped tensor. This data represents static data which is invariant to time change, such as the average junction cardinality or normalized count of categories of that specific region like 'eat, drink and entertainment', 'shopping', 'parking' and 'transport'.

We appended this static info to dynamic input data, so now we have (495, 436, 115) shaped input tensors combined together via the feature channel.

**2.3 Output**

Prediction output is expected to be (6, 495, 436, 8) shaped tensor data. First axis represents the six time bin (future 5, 10, 15, 30, 45, 60 minute) and the last axis represents traffic volume and speed for four headings. In the same fashion as its input, the time axis is combined with the feature channel so (495, 436, 48) tensor data is initially produced as output from our method, then reshaped back to (6, 495, 436, 8) shape for submission.

**2.4 UNet based approaches**

We utilized a UNet [4] based model on Tensorflow [5], having encoder and decoder with skip connections as described in Figure 1. UNet has been widely used in various tasks including image classification and segmentation [9,10,11,12,13]. Mean squared error is used as loss function and evaluation measure, with Adam optimizer [6]. Learning rate started from 3e-4 and manually lowered when loss curve got plateaued while training.

We tried experimenting various net structures, changing composition of layer type and connections between layers. We end up using three model types (named Model 1, 2 and 3), as described in Figure 2 and 3.

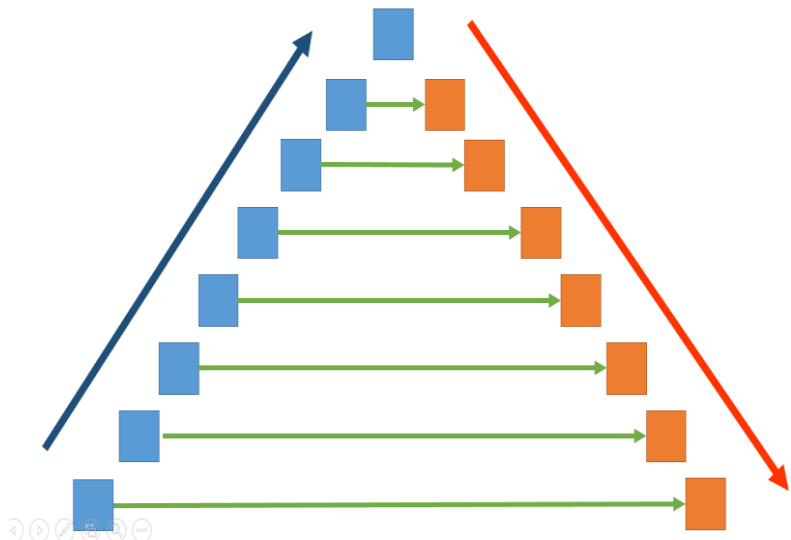

**Figure 1. Overall model structure**
*Each blue box represents dense convolution layers block with average pooling layer. Each orange box represents deconvolution layers. Green arrow represents skip connection between downsampling path and upsamping path.*

*2.4.1 Encoder structure*

Model 1 has similar structure to our last year's experimentation. In each encoding block, each convolution layer is densely connected to every other layer in a feed forward fashion [7], as described in Figure 4.

In Model 2, the max pooling layer is attached in parallel to the dense convolution layers and both their outputs are concatenated before being fed into the last convolution layer. At the end of each block, a convolutional pooling layer is used instead of an average pooling layer.

In Model 3, the max pooling layer is added firsthand, followed by densely connected convolutional layers.

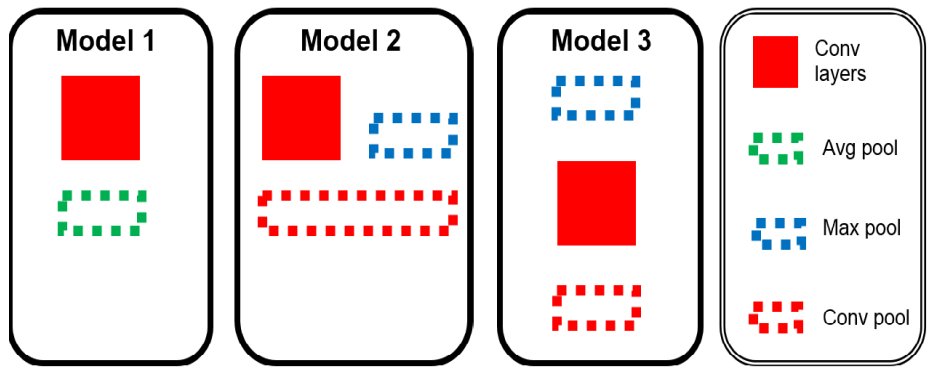

**Figure 2. Encoder structure**

*Each red filled box represents densely connected convolution layers. Green dotted box represents average pooling layer. Blue dotted box represents max pooling layer. Red dotted box represents convolution pooling layer. Every pooling layer's stride size is set to 2.*

*2.4.2 Decoder structure*

In Model 1 and 2, each decoding block has one deconvolution layer, followed by one convolution layer.

In Model 3, linear interpolation layer is attached in parallel to the deconvolution layer and both their outputs are concatenated before being fed into the densely connected convolutional layers.

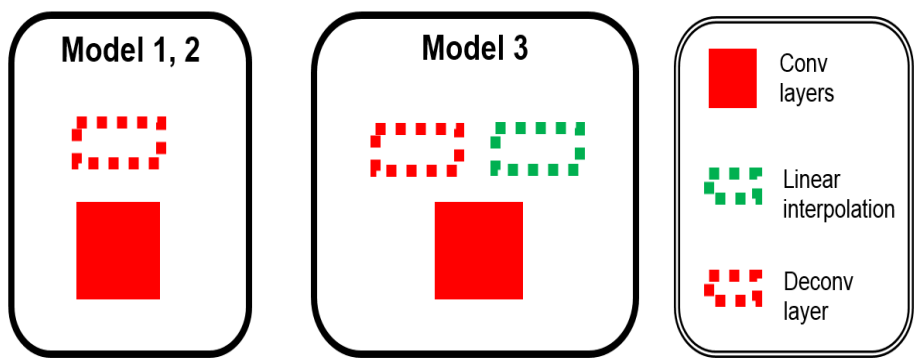

**Figure 3. Decoder structure**

*Each red filled box represents densely connected convolution layers. Green dotted box represents linear interpolation layer. Red dotted box represents deconvolution layer.*

### 2.5 Combining base UNet models prediction

If each model captures different aspects of ground truth that we want to predict, we can expect them to be complementary to improve performance. Meta model aggregate and merge the prediction of multiple base models and produce one final output.

This year we experimented with two simple techniques, which is averaging and taking median [8]. These methods are basically simple math functions, which is easy to implement, and do not require retraining the last layer of the model.

Each base model type has a unique net structure, and even within the same model type performance fluctuates as it undergoes training iterations. We arbitrarily draw multiple base models having various model type and training iteration number, and then combined them.

## 3. Results

Our UNet based model takes input from height 495 width 436 sized input image and incrementally downsized it using average pooling or convolution pooling up to height 4 width 4 image at the top of the UNet (Table 1). Then it regenerates height 495 width 436 sized output image back with prediction. Among three base models we experimented on, there was no clear winner among them. All three individual models showed roughly similar performance, ranging from 1.181 e-3 to 1.169 e-3 in the test set in terms of mean squared error (Table 2). Merging individual models improved performance, test set score ranging from 1.166 e-3 to 1.163 e-3.

**Table 1. Model 1 output shape per each block**

|  | Output shape |
|---|---|
| DenseBlock-1 | (495, 436, 64) |
| AveragePooling | (248, 218, 64) |
| DenseBlock-2 | (248, 218, 96) |
| AveragePooling | (124, 109, 96) |
| DenseBlock-3 | (124, 109, 128) |
| … |  |
| DenseBlock-7 | (8, 7, 128) |
| AveragePooling | (4, 4, 128) |
| DenseBlock-8 | (4, 4, 128) |
| Convolution Layer | (4, 4, 128) |
| DeconvolutionBlock-1 | (8, 7, 128) |
| DeconvolutionBlock-2 | (16, 14, 128) |
| … |  |
| DeconvolutionBlock-7 | (495, 436, 128) |
| Convolution Layer | (495, 436, 48) |

Our best performance 1.1628615 e-3 is achieved by simply averaging six models trained from three base model types (two models drawn per each base model type).

Table 2. Test set evaluation result

|  | Mean squared Error |
|---|---|
| Base Models | 1.169e-3 ~1.181e-3 |
| Merged Models | 1.163e-3 ~1.166e-3 |
| Best performance (6 models merged by averaging) | 1.1628615e-3 |

## 4. Discussions

### 4.1 Base model encoder and decoder structures

Last year we experimented with only Model 1 as a base model. It has a bunch of convolutional layers densely connected to each other in feed forward way (Figure 4) and average pooled when feature map encoded and shrinked in half size to the next level. No max pooling or convolution pooling layer was used there.

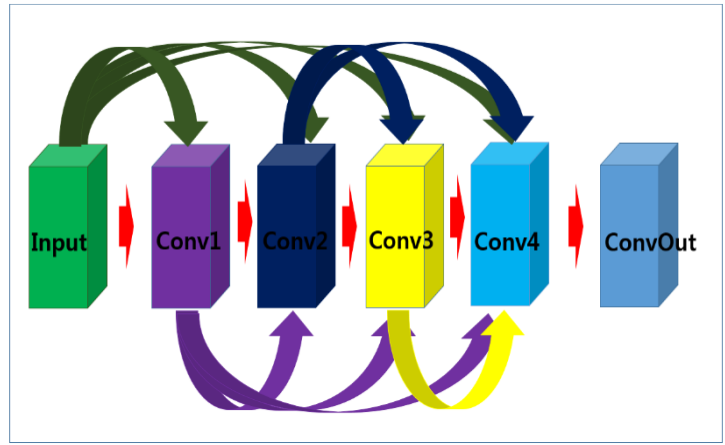

**Figure 4. Dense convolution block structure**
*In our experimentation, convolutional layers are densely connected with each other.*

This year we tried to explore other options possibly useful to make more accurate predictions. In Model 2 and 3, we used a convolution pooling layer instead of average pooling in encoder. In Model 3, in addition to the deconvolution layer, we added a linear interpolation layer in parallel path, which can be understood as a reverse of the average pooling layer. Also, densely connected convolution layers are added to the decoder block.

From evaluation scores from the test set alone, we cannot say that Model 2 or 3 is superior to the Model 1. Performance varies per training iteration but in general there's no significant distinction seen between them in terms of performance. So it's hard to tell we made real improvement by inducing those structural changes. But it could be used as a useful intermediate input to make the final output more accurate.

### 4.2 Static feature vs dynamic feature

Contrary to the previous year's challenge data, this year static data is provided. These static data represents time invariant geographical information. Those information might provide useful clue to the specific region and help build more accurate model. Initially we tried and compared using both dynamic and static data versus having only dynamic data as input feature. But performance gain using static data additionally looks very minimal. We decide to include static data into the input feature anyway because at least it does not hurt performance, but value with regard to the final performance looks very limited in our experimentation.

### 4.3 Benefits of combining multiple models

This year, we tried three UNet base models. Although these base models showed similar performance individually, combined together by averaging, error is reduced further. This suggests possibly each model has a separate, independent error distribution, so averaging output complements each other, making final output more accurate.

Regarding combination method, we only tried two simple mathematical function, which is averaging and taking median. No meaningful performance difference observed between two methods.

## 5. Conclusion

This is our second participation in the Traffic4cast challenge. Starting from last year's experiments and experience, we could further try new ideas with regard to the UNet encoder and decoder's structural compositions. Also we tried to combine base model outputs using simple yet effective methods to improve performance further. Our methods showed effective performance on the real world data collected from three large cities worldwide.